\documentclass[conference]{IEEEtran}
\IEEEoverridecommandlockouts

\usepackage{cite}
\usepackage{amsmath,amssymb,amsfonts}
\usepackage{algorithmic}
\usepackage{graphicx}
\usepackage{textcomp}
\usepackage{xcolor}
\def\BibTeX{{\rm B\kern-.05em{\sc i\kern-.025em b}\kern-.08em
    T\kern-.1667em\lower.7ex\hbox{E}\kern-.125emX}}
\begin{document}

\title{Towards Unified Facial Action Unit Recognition Framework by Large Language Models
}

\author{\IEEEauthorblockN{Guohong Hu, Xing Lan, Hanyu Jiang, Jiayi Lyu and Jian Xue\IEEEauthorrefmark{2}}\thanks{\IEEEauthorrefmark{2}Corresponding Author}\thanks{~Under Review.}
\IEEEauthorblockA{\textit{School of Engineering Science}, \textit{University of Chinese Academy of Sciences}, Beijing, China\\
\{huguogong22,lanxing19,jianghanyu231,lyujiayi21\}@mails.ucas.ac.cn; xuejian@ucas.ac.cn}}

\maketitle


\begin{abstract}

Facial Action Units (AUs) are of great significance in the realm of affective computing. In this paper, we propose AU-LLaVA, the first unified AU recognition framework based on the Large Language Model (LLM). AU-LLaVA consists of a visual encoder, a linear projector layer, and a pre-trained LLM. We meticulously craft the text descriptions and fine-tune the model on various AU datasets, allowing it to generate different formats of AU recognition results for the same input image. On the BP4D and DISFA datasets, AU-LLaVA delivers the most accurate recognition results for nearly half of the AUs. Our model achieves improvements of F1-score up to 11.4\% in specific AU recognition compared to previous benchmark results. On the FEAFA dataset, our method achieves significant improvements over all 24 AUs compared to previous benchmark results. AU-LLaVA demonstrates exceptional performance and versatility in AU recognition.

\end{abstract}


\section{Introduction}
Facial expressions convey crucial nonverbal information. They reveal emotions, intentions, and mental states \cite{bisogni2022impact}, playing a vital role in human communication and social interaction \cite{DARWIN}.
To systematically study facial expressions, Ekman and Friesen developed the Facial Action Coding System (FACS) \cite{FACS} in 1978.
Expression categories and facial Action Units (AUs) are closely related \cite{Du_Tao_Martinez_2014,Friesen_Ekman_1983,Prkachin_1992}.
A reliable AU recognition system is essential for accurate AU detection and intensity estimation. Most of the existing methods rely on Convolutional Neural Network \cite{Li_Abitahi_Zhu_2017}, Graph Neural Network \cite{Chen_Wei_Wang_Guo_2019}, or Transformer \cite{auformer}. Those methods learn AU information encoded in the training set into the backbone, which limits the generalization capability to recognize AU across different datasets. Moreover, they rely solely on visual information and focus exclusively on either detecting the presence of AUs or estimating AU's intensity. Meanwhile, Large Language Models (LLMs) have demonstrated remarkable capabilities in language tasks and reasoning \cite{LLM1,LLM2,LLM3,LLM4}.
Inspired by recent successes in applying LLMs to visual tasks \cite{zhao2024enhancing,kumar2024multimodal,lan2024expllm,wang2024locllm}, we propose AU-LLaVA, a novel approach that leverages LLaVA \cite{liu2024visual} for unified AU recognition.
The proposed method aims to combine the strengths of LLMs with the precision required for facial expression analysis, potentially advancing the field of automated facial action unit recognition.

AU-LLaVA integrates a visual encoder, a linear projector layer, and a pre-trained LLaVA model.
The system utilizes meticulously crafted text descriptions, incorporating task objectives, AU locations and definitions, and expected output formats.
Such text provides the model with a clear problem-oriented focus. When recognizing a specific AU, the model relies on the location description corresponding to which, allowing it to concentrate on a specific region of the facial image, thereby enhancing the model's performance. Moreover, the cues are provided explicitly for the large language model and remain constant during the training and validation.

\begin{figure}[t]
    \centering
    \includegraphics[scale=0.35]{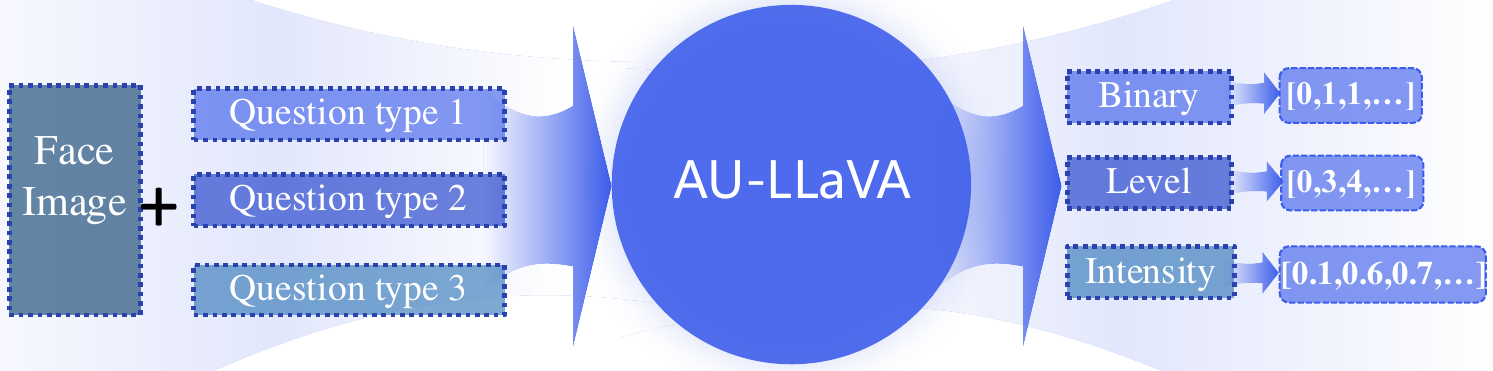} 
    \caption{Versatility of AU-LLaVA results. For a single input image, the model demonstrates multi-modal capabilities: (a) binary AU detection (0 or 1), (b) discrete AU intensity levels (0-5), (c) continuous AU intensity values (0-1). AU-LLaVA is a unified AU recognition framework based on LLM.}
    \label{fig:1} 
\end{figure}

The AU-LLaVA, fine-tuned on diverse AU datasets, demonstrates versatility in generating various AU recognition results for a single input image. 
AU-LLaVA's query-based approach allows for flexible output formats, as illustrated in Fig. \ref{fig:1}. This architecture leverages the strengths of language models in visual tasks, potentially enhancing the accuracy and adaptability of AU recognition systems.
The primary contributions of this study are as follows:
\begin{itemize}
    \item This paper presents AU-LLaVA, the first known unified AU recognition framework based on LLMs. This novel approach integrates visual encoding with the reasoning capabilities of LLMs for facial expression analysis.
    \item  AU-LLaVA demonstrates exceptional efficacy on different tasks, achieving superior F1-score for approximately half of the AUs in BP4D and DISFA datasets. For the FEAFA dataset, AU-LLaVA achieves significant improvements over all 24 AUs compared to previous benchmarks. 
\end{itemize}

\section{Related Work}

\subsection{AU Recognition}\label{AA}
Facial Action Unit (AU) recognition has been extensively researched over the decades, leading to the development of various influential methods.
In previous work, the AU classifier is trained on the extracted features from images.
For example, Baltrusaitis et al.\cite{Baltrusaitis_Mahmoud_Robinson_2015} presented a facial AU intensity estimation and occurrence detection system based on histograms of oriented gradients and geometry features, including landmark locations, which were used to train a Support Vector Machine classifier.
DRML \cite{DRML} addresses region-specific and multi-label learning jointly, while EAC-Net \cite{EAC-Net} employs a single attention map per image, combining all regions associated with action units. 
Yanan Chang et al.\cite{Chang_Wang} proposed a novel knowledge-driven self-supervised representation learning framework for AU recognition. AU labeling rules are summarized and leveraged to guide the framework's design. However, few works leverage the powerful reasoning and generalization capabilities of large language models.

They rely entirely on the model's internal architecture for encoding and reasoning AUs. AUs are learned purely based on visual information.

\subsection{Large Language Model on Vison Tasks }
\label{LLM}

Researchers are integrating LLMs into visual domains to enhance model performance by leveraging additional modalities and contextual information.
Instruction tuning \cite{wei2021finetuned} is widely used to align vision and language modalities, enhancing the capabilities of multimodal LLMs (MLLMs).
In this vein, MiniGPT-4 \cite{zhu2023minigpt} aligns a visual encoder with an advanced LLM, achieving detailed image description generation and other multi-modal capabilities.
Similarly, LLaVA \cite{liu2024visual} leverages GPT-4 generated instruction-following data to create a large multimodal model, demonstrating impressive visual-language understanding and chat abilities.
Although MLLMs have made great advances, most approaches still concentrate on vision-language tasks like Visual Question-Answering (VQA), leaving LLMs' potential in classical vision tasks largely untapped.
In the work related to facial expressions based on large models, EmoLA \cite{li2024facial} addresses the challenges in facial affective behavior analysis by introducing a comprehensive instruction-following dataset and benchmark. It enhances MLLM's performance by incorporating a facial prior expert module with face structure knowledge and employing a low-rank adaptation module for efficient fine-tuning. EMO-LLaMA \cite{xing2024emo} incorporates facial priors through a Face Info Mining module and leverages handcrafted prompts with age-gender-race attributes, enhancing the model's ability to  interpret facial expressions across diverse human groups.

\begin{figure}[t]
    \centering
    \includegraphics[width=0.90\linewidth]{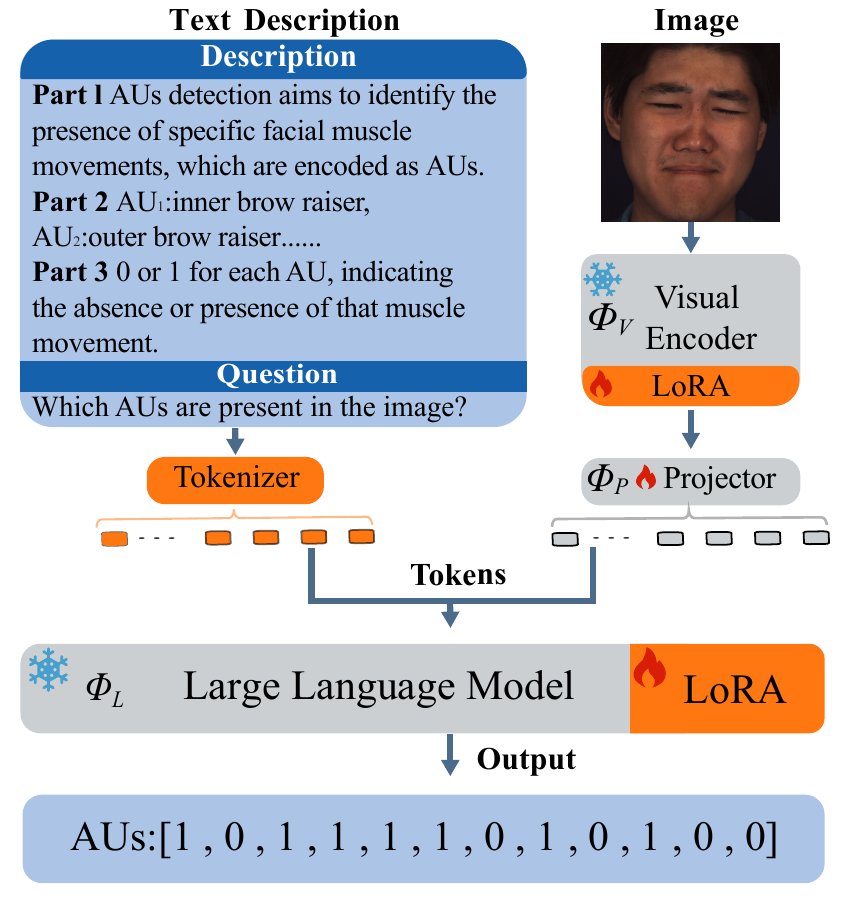} 
    \caption{Architectural framework of AU-LLaVA, which comprises three primary components: a visual encode, a linear projector, and a pretrained LLM. AU-LLaVA processes facial images and textual descriptions as inputs, generating an array where each element corresponds to a specific Action Unit. During the training phase, Low-Rank Adaptation (LoRA) modules are integrated into both the visual encoder and the LLM to enhance efficiency.}
    \label{fig:2} 
\end{figure}

\section{The Proposed Method}
\label{sec:method}
\subsection{Overview}
This paper leverages the powerful capabilities of a LLM for reasoning, employing the text description and AU datasets with various formats of AU labels to fine-tune LLM. We define LLM-based AU recognition as a VQA task and propose AU-LLaVA. 
As shown in \eqref{eq:input-output}, given an input facial image $\mathbf{I}$ and the text description $\mathbf{T}$, AU-LLaVA aims to output an array $\mathbf{A}$ where each element corresponds to a type of AU. 
\begin{equation}
\label{eq:input-output} 
\mathbf{A}^{i} =\operatorname{recognize}\left( \mathbf{I},\mathbf{T}^{i}  \right),      
\end{equation}
where $i$ represents the index of three types of AU recognition tasks with different output formats:
\begin{enumerate}
    \item  Integers in $\{0, 1\}$ to indicate the AU's presence or absence.
    \item  Integers in $[0,5]$ to represent the level of an AU.
    \item  Floating-point numbers in $[0,1]$ to represent the intensity of an AU.
\end{enumerate}

AU-LLaVA adopts a similar architecture to LLaVA \cite{liu2024visual}. As shown in Fig. \ref{fig:2}, AU-LLaVA consists three main components: a visual encoder $\Phi_{V}$, a linear projector $\Phi_{P}$ and a large language model $\Phi_{L}$.
 $\Phi_{V}$ takes facial image $\mathbf{I}$ as input and outputs a series of image features, which are then passed to $\Phi_{P}$ to generate a sequence of image tokens. 
 The text description $\mathbf{T}$ is transformed into a sequence of text tokens through a tokenizer. Subsequently, the text tokens and image tokens are combined and fed into the $\Phi_{L}$.
 This allows process Eq. \ref{eq:input-output}  to be further specified as Eq. \ref{eq: whole process}. 
\begin{equation}
\label{eq: whole process} 
\mathbf{A}^{i}= \Phi_{L} \left( \Phi_{P} \left( \Phi_{V} \left( \mathbf{I} \right)  \right), \mathbf{T}^{i}  \right).
\end{equation}

\subsection{Text Description}
It is shown that the textual instruction $\mathbf{T}$ plays a crucial role in manifesting the capabilities of LLMs \cite{Xuan_Guo_Yang_Zhang_2023}. When designing $\mathbf{T}$, similar to previous VQA tasks, we assign tailored questions to each task. Additionally, we meticulously construct descriptions comprising three essential elements: Part 1 is the purpose of the current AU recognition task, which aims to inform the model of what specific task it needs to perform.    
For instance, when the model is instructed in an AU detection task, Part 1 of $\mathbf{T}$ directs the model's focus solely on the presence or absence of AUs without considering the intensity of each AU, providing a clear goal-oriented orientation and improving efficiency.
Part 2 introduces the AUs, specifying that each AU represents a specific facial muscle action. 
Such an introduction helps the model establish a fundamental understanding and cognitive framework for AUs. Additionally, it indicates which facial regions the LLM should focus on when recognizing a specific AU. Part 3 is the description of the expected output format from the model. For example, in an AU detection task, the model is expected to output either 0 or 1 for each AU, indicating whether the AU is present or not. It clarifies the model's output requirements, ensuring that the results align with the task's objectives. A sample of a detailed description of textual description $\mathbf{T}$ can be found in Fig. \ref{fig:2}.
After being processed by a tokenizer, the entire textual description $\mathbf{T}$ is converted into text tokens, which are then fed into a LLM $\Phi _{L}$ for further processing and analyzing.

\subsection{LoRA Modules}
To uncover and enhance the performance of LLM on specific recognition tasks, it is necessary to fine-tune the pre-trained LLM using AU datasets. However, given the immense number of parameters in LLM, it is impractical to train the entire model's parameters from scratch with limited GPU and a finite dataset size. Inspired by the work of Hu et al.\cite{J._Shen_Wallis_Allen-Zhu_Li_Wang_Chen_2021}, the LoRA technique is employed to the training of AU-LLaVA.  LoRA is a parameter-efficient fine-tuning technique that specializes pre-trained models by introducing trainable low-rank matrices at each layer, significantly reducing the number of trainable parameters. The entire fine-tuning process in LoRA can be mathematically expressed as follows:
\begin{equation}
\label{eq:LoRA}
\mathbf{W}=\mathbf{W}_{0}+\mathbf{U}\times\mathbf{V} 
\end{equation}
$\mathbf{W}_{0}$ represents a pre-trained weight matrix, whereas $\mathbf{W}$ denotes the fine-tuned weight matrix. $\mathbf{U}\in \mathbb{R}^{d\times r} $ and $\mathbf{V}\in \mathbb{R}^{r\times k} $ signify two low-rank matrices that satisfy the condition $r\ll k$. During the training procedure, freeze $\mathbf{W}_{0}$, and only $\mathbf{U}$ and $\mathbf{V}$ are updated. Consequently, only the the parameters within $\mathbf{U}$ and $\mathbf{V}$ need to be trained, significantly reducing the number of trainable parameters compared to fine-tuning the entire model. Specifically, with regard to our AU-LLaVA, we employ LoRA modules into both visual encoder $\Phi_{V}$ and LLM $\Phi_{L}$, and fine-tune them with the linear projector.

\section{Experiments}
\label{sec:experiments}
\subsection{Datasets}
\textbf{BP4D} \cite{Zhang_Yin_Cohn_Canavan_Reale_Horowitz_Liu_Girard_2014} dataset is widely used in AU detection. It contains 41 subjects with 23 females and 18 males, each of which is involved in 8 different tasks. It includes approximately 140,000 face images with binary AU labels ($1$ for presence or $0$ for absence). 
\textbf{DISFA}\cite{Mavadati_Mahoor_Bartlett_Trinh_Cohn_2013} dataset contains 27 subjects with 12 females and 15 males. It includes approximately 138,000 face images annotated with AU intensities (discrete level $0$ to $5$). 
\textbf{FEAFA}\cite{FEAFA,FEAFA+} dataset features 122 participants in naturalistic settings. In addition, 99,356 frames are manually labeled using continuous AU intensities (in the continuous range $[0,1]$).

\subsection{Implementation Details}
\label{sseq:settings}
Face detection and alignment were performed using SCRFD \cite{scrfd} on each dataset. Before feeding the facial images into AU-LLaVA, they were cropped to a size of 224$\times$224. Adhering to the protocol established in previous works\cite{2017Action,2019Local}, subject-independent 3-fold cross-validation was conducted on BP4D and DISFA. The average results across these 3 folds were then reported. For FEAFA dataset, we randomly split it into training and validation subsets with a ratio of 4:1. 
We utilize ViT-L/14 as the visual encoder $\Phi_{V}$, which is pre-trained with DINOv2 \cite{Dinov2} weights. We employ an instruction-tuned Vicuna7B \cite{vicuna} as the pre-trained large language model $\Phi_{L}$. The projector $\Phi_{P}$ consists of a single fully connected layer. 

During training, data augmentation techniques were applied. The AdamW optimizer was utilized with a batch size of 1 per GPU, and gradient accumulation was performed over 16 steps. 
The learning rate was initially set to 5e-4 and followed a cosine decay schedule, with a weight decay of 0.05 and a warm-up ratio of 0.03. In the first stage, the model was trained for 5 epochs, with LoRA fine-tuning applied to each self-attention layer of the visual encoder and the LLM, using a hidden dimension $r$ of 8. AU-LLaVA was then trained for approximately 4 days on each of the three datasets using four V100 GPUs.

\subsection{Ablation Study}



Ablation experiments were conducted on the BP4D dataset, following the experimental setup outlined in Section \ref{sseq:settings}. The fine-tuning procedure was modified to fine-tune exclusively on either the LLM $\Phi_{L}$ or the visual encoder $\Phi_{V}$. The results are shown in Table \ref{tab:Ablation}, where $\Phi_{L}$ \& $\Phi_{V}$ represents the full fine-tuning strategy of AU-LLaVA. Four AUs were selected for comparison, as they correspond to the eye and mouth regions of the face, respectively. The ablation study confirms that the designed training strategies lead to enhanced performance.

\begin{table}[htb]
\centering
\caption{Ablation study on fine-tuning each component of AU-LLaVA on BP4D, using the F1-score metric(in \%).}\label{tab:Ablation}
\begin{tabular}{cccccc}
\hline
FT Comp.     & AU2                 & AU4                 & AU12                & AU17                & \multicolumn{1}{c}{Avg.}                \\ \hline
$\Phi _{V}$             & 56.0                & 57.1                & 89.3                & 61.8                & 58.4                                    \\
$\Phi _{L}$             & 52.1                & 43.6                & {\underline{\textbf{90.5}}} & 56.7                & 53.7                                    \\ \hline
$\Phi _{L}$ \& $\Phi _{V}$ & {\underline{\textbf{58.2}}} & {\underline{\textbf{61.9}}} & {\underline{\textbf{90.5}}} & {\underline{\textbf{62.5}}} & \multicolumn{1}{c}{{\underline{\textbf{60.3}}}} \\ \hline
\end{tabular}
\end{table}

\subsection{Comparison with state-of-the-art Methods}
\begin{table*}[htb]
\caption{Comparison with the related methods on BP4D dataset using the F1 score metric (in \%). The numbers bolded and underlined represent the best performance.}\label{tab:BP4D}
\centering
\begin{tabular}{cccccccccccccc}
\hline
Method   & AU1                 & AU2                 & AU4                 & AU6                 & AU7                 & AU10                & AU12                & AU14                & AU15                & AU17                & AU23                & AU24                & Avg.                \\ \hline
LSVM\cite{LSVM}     & 23.2                & 22.8                & 23.1                & 27.2                & 47.1                & 77.2                & 63.7                & \underline{\textbf{64.3}} & 18.4                & 33.0                & 19.4                & 20.7                & 35.3                \\
DRML\cite{DRML}     & 36.4                & 41.8                & 43.0                & 55.0                & 67.0                & 66.3                & 65.8                & 54.1                & 33.2                & 48.0                & 31.7                & 30.0                & 48.3                \\
DSIN\cite{DSIN}     & 51.7                & 40.4                & 56.0                & 76.1                & 73.5                & 79.9                & 85.4                & 62.7                & 37.3                & 62.9                & 38.8                & 41.6                & 58.9                \\
EAC\cite{EAC-Net}      & 39.0                & 35.2                & 48.6                & 76.1                & 72.9                & 81.9                & 86.2                & 58.8                & 37.5                & 59.1                & 35.9                & 35.8                & 55.9                \\

SRERL\cite{SRERL}    & 46.9                & 45.3                & 55.6                & 77.1                & \underline{\textbf{78.4}} & 83.5                & 87.6                & 53.9                & \underline{\textbf{53.2}} & \underline{\textbf{63.9}} & \underline{\textbf{47.1}}               & \underline{\textbf{53.3}}                & 62.1                \\
JAA\cite{JAA_old}      & 53.8                & 47.8                & 58.2                & 78.5                & 75.8                & 82.7                & 88.2                & 63.7                & 43.3                & 61.8                & 45.6                & 49.9                & \underline{\textbf{62.4}} \\ \hline
\textbf{AU-LLaVA (Ours)} & \underline{\textbf{58.2}} & \underline{\textbf{45.9}} & \underline{\textbf{61.9}} & \underline{\textbf{78.6}} & 75.6                & \underline{\textbf{87.8}} & \underline{\textbf{90.5}} & 59.0                & 32.4                & 62.5                & 30.5                & 40.3                & 60.3                \\ \hline
\end{tabular}
\end{table*}

\begin{table*}[]
\centering
\caption{Comparison with the related methods on DISFA dataset using the F1 score metric (in \%). The numbers bolded and underlined represent the best performance.}\label{tab:DISFA}
\begin{tabular}{cccccccccc}
\hline
Method   & AU1                      & AU2                      & AU4                      & AU6                      & AU9                      & AU12                     & AU25                     & AU26                                    & Avg.                                    \\ \hline
LSVM\cite{LSVM}     & 10.8                     & 10.0                     & 21.8                     & 15.7                     & 11.5                     & 70.4                     & 12.0                     & 22.1                                    & 21.8                                    \\
DRML\cite{DRML}     & 17.3                     & 17.7                     & 37.4                     & 29.0                     & 10.7                     & 37.7                     & 38.5                     & 20.1                                    & 26.7                                    \\
DSIN\cite{DSIN}     & 42.4                     & 39.0                     & 68.4                     & 28.6                     & 46.8                     & 70.8                     & 90.4                     & 42.2                                    & 53.6                                    \\
EAC\cite{EAC-Net}      & 41.5                     & 26.4                     & 66.4                     & \underline{\textbf{50.7}}      & \underline{\textbf{80.5}}      & \underline{\textbf{89.3}}      & 88.9                     & 15.6                                    & 48.5                                    \\
SRERL\cite{SRERL}    & 45.7                     & 47.8                     & 59.6                     & 47.1                     & 45.6                     & 73.5                     & 84.3                     & 43.6                                    & 55.9                                    \\
JAA\cite{JAA_old}      & 43.7 & 46.2 & 56.0 & 41.4 & 44.7 & 69.6 & 88.3 & \underline{\textbf{58.4}} & \underline{\textbf{56.0}} \\ \hline
\textbf{AU-LLaVA (Ours)} & \underline{\textbf{52.0}}      & \underline{\textbf{59.2}}      & 44.4                     & 30.8                     & 22.3                     & 66.1                     & \underline{\textbf{90.8}}      & 54.6                                    & 52.5                                    \\ \hline
\end{tabular}
\end{table*}

\begin{table*}[htb]
\caption{Comparison with related methods on FEAFA Dataset using the MAE (lower is better) metric. The numbers bolded and underlined represent the best performance.}\label{tab:FEAFA}
\centering
\begin{tabular}{cccccccccccccc}
\hline
Method   & AU1                 & AU3                 & AU5                 & AU7                 & AU9                 & AU11                & AU13                & AU15                & AU17                & AU19                & AU21                & AU23                & Avg.                \\ \hline
DRML\cite{DRML}     & .156                & .052                & .102                & .063                & .045                & .022                & .066                & .054                & .053                & .056                & .045                & .040                & .549                \\
JAA\cite{JAA_old}      & .128                & .042                & .085                & .059                & .045                & .022                & .067                & .046                & .045                & .048                & .045                & .033                & .056                \\
RA-UWML\cite{RA-UWML}  & .098                & .041                & .058                & .051                & .043                & .022                & .062                & .050                & .038                & .045                & .043                & .029                & .048                \\ \hline
\textbf{AU-LLaVA (Ours)} & {\underline{\textbf{.024}}} & {\underline{\textbf{.006}}} & {\underline{\textbf{.012}}} & {\underline{\textbf{.011}}} & {\underline{\textbf{.010}}} & {\underline{\textbf{.007}}} & {\underline{\textbf{.014}}} & {\underline{\textbf{.012}}} & {\underline{\textbf{.012}}} & {\underline{\textbf{.010}}} & {\underline{\textbf{.008}}} & {\underline{\textbf{.005}}} & {\underline{\textbf{.011}}} \\ \hline
\end{tabular}
\end{table*}

We compared our method with current AU recognition work. The results for other methods are sourced directly from their respective publications.
The F1-score is calculated for 12 AUs in BP4D and 8 AUs in DISFA across the three cross-validation folds. The proposed method is compared with existing AU recognition approaches based on F1-score. 
Table \ref{tab:BP4D} shows the performance comparison of AU-LLaVA with other AU recognition methods on BP4D. Our method achieves the best accuracy in recognizing AU1, AU2, AU4, AU6, AU10, and AU12. 
Table \ref{tab:DISFA} shows the comparison on DISFA. Our method achieves the best accuracy in recognizing AU1, AU2, and AU25. 
The F1-score is improved for nearly half of the AUs across both datasets. Moreover, AU-LLaVA achieves significant improvements in the recognition of certain AUs. For instance, on the BP4D dataset, AU1, AU4, and AU10 exhibit an average improvement of 4.13 percentage points. Meanwhile, On the DISFA dataset, AU-LLaVA significantly surpasses the previous best method in recognizing AU2, with an improvement of 11.4 percentage points. 


Notably, when training on the DISFA dataset, AU-LLaVA is expected to output levels of AUs (from 0 to 5), thus we used the original dataset labels for training. 
During validation, levels 2 and below were treated as 0, and levels above 2 as 1, which led to a slight decrease in performance compared to the BP4D dataset.
Additionally, the previous methods compared with AU-LLaVA rely solely on visual information, whereas AU-LLaVA integrates both visual and textual modalities and captures correlations through cross-modality attention.
Moreover, we do not provide a predefined relationship between AUs, even though the occurrence of one AU is often linked to others. If these relationships were explicitly defined before training, as done in SRERL \cite{SRERL}, further improvements in performance could be anticipated.


For the FEAFA dataset, since the AU labels are continuous intensity values between 0 and 1, the Mean Absolute Error (MAE) is computed for the 24 AUs. Table \ref{tab:FEAFA} presents the comparison for 13 AUs along with the average performance. In fact, AU-LLaVA achieved significant improvements over the previous best results across all 24 AUs. Note that both JAA \cite{JAA_old} and DRML \cite{DRML} are proposed for AU detection rather than AU intensity estimation. When adapting them for intensity estimation, their performance is not as strong on the BP4D and DISFA datasets. The results demonstrate the strong generalization ability of our model. When adapting AU-LLaVA to different tasks, the only change required is the text description.

\section{Conclusion}
This paper introduces AU-LLaVA, a pioneering unified framework for facial Action Unit (AU) recognition leveraging large language models (LLMs). By harnessing the advanced reasoning capabilities of LLMs, AU-LLaVA effectively recognizes AUs across diverse tasks. This approach not only demonstrates the potential of integrating textual and visual modalities but also sets a new standard for facial AU recognition.
Potential future research directions include investigating the impact of integrating AU relationships into the text descriptions, as well as applying AU-LLaVA to other visual tasks such as recognizing facial micro-expressions and detecting facial landmarks.

\bibliographystyle{IEEEtran}
\bibliography{IEEEexample}

\begin{thebibliography}{10}
\providecommand{\url}[1]{#1}
\csname url@samestyle\endcsname
\providecommand{\newblock}{\relax}
\providecommand{\bibinfo}[2]{#2}
\providecommand{\BIBentrySTDinterwordspacing}{\spaceskip=0pt\relax}
\providecommand{\BIBentryALTinterwordstretchfactor}{4}
\providecommand{\BIBentryALTinterwordspacing}{\spaceskip=\fontdimen2\font plus
\BIBentryALTinterwordstretchfactor\fontdimen3\font minus \fontdimen4\font\relax}
\providecommand{\BIBforeignlanguage}[2]{{%
\expandafter\ifx\csname l@#1\endcsname\relax
\typeout{** WARNING: IEEEtran.bst: No hyphenation pattern has been}%
\typeout{** loaded for the language `#1'. Using the pattern for}%
\typeout{** the default language instead.}%
\else
\language=\csname l@#1\endcsname
\fi
#2}}
\providecommand{\BIBdecl}{\relax}
\BIBdecl

\bibitem{bisogni2022impact}
C.~Bisogni, A.~Castiglione, S.~Hossain, F.~Narducci, and S.~Umer, ``Impact of deep learning approaches on facial expression recognition in healthcare industries,'' \emph{IEEE Transactions on Industrial Informatics}, vol.~18, no.~8, pp. 5619--5627, 2022.

\bibitem{DARWIN}
\BIBentryALTinterwordspacing
C.~DARWIN, ``\BIBforeignlanguage{en-US}{The expression of the emotions in man and animals},'' \emph{\BIBforeignlanguage{en-US}{The American Journal of the Medical Sciences}}, p. 477. [Online]. Available: \url{http://dx.doi.org/10.1097/00000441-195610000-00024}
\BIBentrySTDinterwordspacing

\bibitem{FACS}
P.~Ekman and W.~V. Friesen, ``Facial action coding system (facs): a technique for the measurement of facial actions,'' \emph{Rivista Di Psichiatria}, vol.~47, no.~2, pp. 126--38, 1978.

\bibitem{Du_Tao_Martinez_2014}
\BIBentryALTinterwordspacing
S.~Du, Y.~Tao, and A.~M. Martinez, ``\BIBforeignlanguage{en-US}{Compound facial expressions of emotion.}'' \emph{\BIBforeignlanguage{en-US}{Proceedings of the National Academy of Sciences}}, Apr 2014. [Online]. Available: \url{http://dx.doi.org/10.1073/pnas.1322355111}
\BIBentrySTDinterwordspacing

\bibitem{Friesen_Ekman_1983}
W.~Friesen and P.~Ekman, ``\BIBforeignlanguage{en-US}{Emfacs-7: Emotional facial action coding system},'' Jan 1983.

\bibitem{Prkachin_1992}
\BIBentryALTinterwordspacing
K.~M. Prkachin, ``\BIBforeignlanguage{en-US}{The consistency of facial expressions of pain: a comparison across modalities},'' \emph{\BIBforeignlanguage{en-US}{Pain}}, vol.~51, no.~3, p. 297–306, Dec 1992. [Online]. Available: \url{http://dx.doi.org/10.1016/0304-3959(92)90213-u}
\BIBentrySTDinterwordspacing

\bibitem{Li_Abitahi_Zhu_2017}
W.~Li, F.~Abitahi, and Z.~Zhu, ``\BIBforeignlanguage{en-US}{Action unit detection with region adaptation, multi-labeling learning and optimal temporal fusing},'' \emph{\BIBforeignlanguage{en-US}{Cornell University - arXiv,Cornell University - arXiv}}, Apr 2017.

\bibitem{Chen_Wei_Wang_Guo_2019}
\BIBentryALTinterwordspacing
Z.-M. Chen, X.-S. Wei, P.~Wang, and Y.~Guo, ``\BIBforeignlanguage{en-US}{Multi-label image recognition with graph convolutional networks},'' in \emph{\BIBforeignlanguage{en-US}{2019 IEEE/CVF Conference on Computer Vision and Pattern Recognition (CVPR)}}, Jun 2019. [Online]. Available: \url{http://dx.doi.org/10.1109/cvpr.2019.00532}
\BIBentrySTDinterwordspacing

\bibitem{auformer}
K.~Yuan, Z.~Yu, X.~Liu, W.~Xie, H.~Yue, and J.~Yang, ``Auformer: Vision transformers are parameter-efficient facial action unit detectors,'' \emph{arXiv preprint arXiv:2403.04697}, 2024.

\bibitem{LLM1}
L.~Ouyang, J.~Wu, X.~Jiang \emph{et~al.}, ``\BIBforeignlanguage{en-US}{Training language models to follow instructions with human feedback}.''

\bibitem{LLM2}
J.~Hoffmann, S.~Borgeaud, A.~Mensch, E.~Buchatskaya \emph{et~al.}, ``\BIBforeignlanguage{en-US}{Training compute-optimal large language models}.''

\bibitem{LLM3}
T.~Brown, B.~Mann, N.~Ryder, M.~Subbiah, J.~Kaplan, P.~Dhariwal,  \emph{et~al.}, ``\BIBforeignlanguage{en-US}{Language models are few-shot learners},'' \emph{\BIBforeignlanguage{en-US}{arXiv: Computation and Language,arXiv: Computation and Language}}, May 2020.

\bibitem{LLM4}
J.~Li, D.~Li, S.~Savarese, and S.~Hoi, ``Blip-2: Bootstrapping language-image pre-training with frozen image encoders and large language models,'' in \emph{International conference on machine learning}.\hskip 1em plus 0.5em minus 0.4em\relax PMLR, 2023, pp. 19\,730--19\,742.

\bibitem{zhao2024enhancing}
Z.~Zhao, Y.~Cao, S.~Gong, and I.~Patras, ``Enhancing zero-shot facial expression recognition by llm knowledge transfer,'' \emph{arXiv preprint arXiv:2405.19100}, 2024.

\bibitem{kumar2024multimodal}
C.~O. Kumar, N.~Gowtham, M.~Zakariah, and A.~Almazyad, ``Multimodal emotion recognition using feature fusion: An llm-based approach,'' \emph{IEEE Access}, 2024.

\bibitem{lan2024expllm}
X.~Lan, J.~Xue, J.~Qi, D.~Jiang, K.~Lu, and T.-S. Chua, ``Expllm: Towards chain of thought for facial expression recognition,'' \emph{arXiv preprint arXiv:2409.02828}, 2024.

\bibitem{wang2024locllm}
D.~Wang, S.~Xuan, and S.~Zhang, ``Locllm: Exploiting generalizable human keypoint localization via large language model,'' in \emph{Proceedings of the IEEE/CVF Conference on Computer Vision and Pattern Recognition}, 2024, pp. 614--623.

\bibitem{liu2024visual}
H.~Liu, C.~Li, Q.~Wu, and Y.~J. Lee, ``Visual instruction tuning,'' \emph{Advances in neural information processing systems}, vol.~36, 2024.

\bibitem{Baltrusaitis_Mahmoud_Robinson_2015}
\BIBentryALTinterwordspacing
T.~Baltrusaitis, M.~Mahmoud, and P.~Robinson, ``\BIBforeignlanguage{en-US}{Cross-dataset learning and person-specific normalisation for automatic action unit detection},'' in \emph{\BIBforeignlanguage{en-US}{2015 11th IEEE International Conference and Workshops on Automatic Face and Gesture Recognition (FG)}}, May 2015. [Online]. Available: \url{http://dx.doi.org/10.1109/fg.2015.7284869}
\BIBentrySTDinterwordspacing

\bibitem{DRML}
\BIBentryALTinterwordspacing
K.~Zhao, W.-S. Chu, and H.~Zhang, ``\BIBforeignlanguage{en-US}{Deep region and multi-label learning for facial action unit detection},'' in \emph{\BIBforeignlanguage{en-US}{2016 IEEE Conference on Computer Vision and Pattern Recognition (CVPR)}}, Jun 2016. [Online]. Available: \url{http://dx.doi.org/10.1109/cvpr.2016.369}
\BIBentrySTDinterwordspacing

\bibitem{EAC-Net}
\BIBentryALTinterwordspacing
W.~Li, F.~Abtahi, Z.~Zhu, and L.~Yin, ``\BIBforeignlanguage{en-US}{Eac-net: Deep nets with enhancing and cropping for facial action unit detection},'' \emph{\BIBforeignlanguage{en-US}{IEEE Transactions on Pattern Analysis and Machine Intelligence}}, p. 2583–2596, Nov 2018. [Online]. Available: \url{http://dx.doi.org/10.1109/tpami.2018.2791608}
\BIBentrySTDinterwordspacing

\bibitem{Chang_Wang}
Y.~Chang and S.~Wang, ``\BIBforeignlanguage{en-US}{Knowledge-driven self-supervised representation learning for facial action unit recognition}.''

\bibitem{wei2021finetuned}
J.~Wei, M.~Bosma, V.~Y. Zhao, K.~Guu, A.~W. Yu, B.~Lester, N.~Du, A.~M. Dai, and Q.~V. Le, ``Finetuned language models are zero-shot learners,'' \emph{arXiv preprint arXiv:2109.01652}, 2021.

\bibitem{zhu2023minigpt}
D.~Zhu, J.~Chen, X.~Shen, X.~Li, and M.~Elhoseiny, ``Minigpt-4: Enhancing vision-language understanding with advanced large language models,'' \emph{arXiv preprint arXiv:2304.10592}, 2023.

\bibitem{li2024facial}
Y.~Li, A.~Dao, W.~Bao, Z.~Tan, T.~Chen, H.~Liu, and Y.~Kong, ``Facial affective behavior analysis with instruction tuning,'' \emph{arXiv preprint arXiv:2404.05052}, 2024.

\bibitem{xing2024emo}
B.~Xing, Z.~Yu, X.~Liu, K.~Yuan, Q.~Ye, W.~Xie, H.~Yue, J.~Yang, and H.~K{\"a}lvi{\"a}inen, ``Emo-llama: Enhancing facial emotion understanding with instruction tuning,'' \emph{arXiv preprint arXiv:2408.11424}, 2024.

\bibitem{Xuan_Guo_Yang_Zhang_2023}
S.~Xuan, Q.~Guo, M.~Yang, and S.~Zhang, ``\BIBforeignlanguage{en-US}{Pink: Unveiling the power of referential comprehension for multi-modal llms},'' Oct 2023.

\bibitem{J._Shen_Wallis_Allen-Zhu_Li_Wang_Chen_2021}
H.~J., Y.~Shen, P.~Wallis, Z.~Allen-Zhu, Y.~Li, S.~Wang, and W.~Chen, ``\BIBforeignlanguage{en-US}{Lora: Low-rank adaptation of large language models.}'' \emph{\BIBforeignlanguage{en-US}{arXiv: Computation and Language,arXiv: Computation and Language}}, Jun 2021.

\bibitem{Zhang_Yin_Cohn_Canavan_Reale_Horowitz_Liu_Girard_2014}
\BIBentryALTinterwordspacing
X.~Zhang, L.~Yin, J.~F. Cohn, S.~Canavan, M.~Reale, A.~Horowitz, P.~Liu, and J.~M. Girard, ``\BIBforeignlanguage{en-US}{Bp4d-spontaneous: a high-resolution spontaneous 3d dynamic facial expression database},'' \emph{\BIBforeignlanguage{en-US}{Image and Vision Computing}}, p. 692–706, Oct 2014. [Online]. Available: \url{http://dx.doi.org/10.1016/j.imavis.2014.06.002}
\BIBentrySTDinterwordspacing

\bibitem{Mavadati_Mahoor_Bartlett_Trinh_Cohn_2013}
\BIBentryALTinterwordspacing
S.~M. Mavadati, M.~H. Mahoor, K.~Bartlett, P.~Trinh, and J.~F. Cohn, ``\BIBforeignlanguage{en-US}{Disfa: A spontaneous facial action intensity database},'' \emph{\BIBforeignlanguage{en-US}{IEEE Transactions on Affective Computing}}, p. 151–160, Apr 2013. [Online]. Available: \url{http://dx.doi.org/10.1109/t-affc.2013.4}
\BIBentrySTDinterwordspacing

\bibitem{FEAFA}
Y.~Yan, K.~Lu, J.~Xue, P.~Gao, and J.~Lyu, ``\BIBforeignlanguage{en-US}{Feafa: A well-annotated dataset for facial expression analysis and 3d facial animation},'' \emph{\BIBforeignlanguage{en-US}{Cornell University - arXiv,Cornell University - arXiv}}, Apr 2019.

\bibitem{FEAFA+}
W.~Gan, J.~Xue, K.~Lu, Y.~Yan, P.~Gao, and J.~Lyu, ``Feafa+: an extended well-annotated dataset for facial expression analysis and 3d facial animation,'' in \emph{Fourteenth International Conference on Digital Image Processing (ICDIP 2022)}, vol. 12342.\hskip 1em plus 0.5em minus 0.4em\relax SPIE, 2022, pp. 307--316.

\bibitem{scrfd}
\BIBentryALTinterwordspacing
J.~Guo, J.~Deng, A.~Lattas, and S.~Zafeiriou, ``Sample and computation redistribution for efficient face detection,'' 2021. [Online]. Available: \url{https://arxiv.org/abs/2105.04714}
\BIBentrySTDinterwordspacing

\bibitem{2017Action}
W.~Li, F.~Abitahi, and Z.~Zhu, ``Action unit detection with region adaptation, multi-labeling learning and optimal temporal fusing,'' \emph{IEEE}, 2017.

\bibitem{2019Local}
X.~Niu, H.~Han, S.~Yang, Y.~Huang, and S.~Shan, ``Local relationship learning with person-specific shape regularization for facial action unit detection,'' \emph{IEEE}, 2019.

\bibitem{Dinov2}
M.~Oquab, T.~Darcet, T.~Moutakanni, H.~Vo, M.~Szafraniec, V.~Khalidov, P.~Fernandez, D.~Haziza, F.~Massa, A.~El-Nouby \emph{et~al.}, ``Dinov2: Learning robust visual features without supervision,'' \emph{arXiv preprint arXiv:2304.07193}, 2023.

\bibitem{vicuna}
\BIBentryALTinterwordspacing
W.-L. Chiang, Z.~Li, Z.~Lin \emph{et~al.}, ``Vicuna: An open-source chatbot impressing gpt-4 with 90\%* chatgpt quality,'' March 2023. [Online]. Available: \url{https://lmsys.org/blog/2023-03-30-vicuna/}
\BIBentrySTDinterwordspacing

\bibitem{LSVM}
F.-E. FanRong-En, C.-W. ChangKai-Wei, H.-J. HsiehCho-Jui, W.-R. WangXiang-Rui, and L.-J. LinChih-Jen, ``\BIBforeignlanguage{en-US}{Liblinear: A library for large linear classification},'' \emph{\BIBforeignlanguage{en-US}{Journal of Machine Learning Research,Journal of Machine Learning Research}}, Jun 2008.

\bibitem{DSIN}
C.~Corneanu, M.~Madadi, and S.~Escalera, ``\BIBforeignlanguage{en-US}{Deep structure inference network for facial action unit recognition},'' \emph{\BIBforeignlanguage{en-US}{Cornell University - arXiv,Cornell University - arXiv}}, Mar 2018.

\bibitem{SRERL}
\BIBentryALTinterwordspacing
G.~Li, X.~Zhu, Y.~Zeng, Q.~Wang, and L.~Lin, ``\BIBforeignlanguage{en-US}{Semantic relationships guided representation learning for facial action unit recognition},'' \emph{\BIBforeignlanguage{en-US}{Proceedings of the AAAI Conference on Artificial Intelligence}}, p. 8594–8601, Aug 2019. [Online]. Available: \url{http://dx.doi.org/10.1609/aaai.v33i01.33018594}
\BIBentrySTDinterwordspacing

\bibitem{JAA_old}
Z.~Shao, Z.~Liu, J.~Cai, and L.~Ma, ``Deep adaptive attention for joint facial action unit detection and face alignment,'' 2018.

\bibitem{RA-UWML}
C.~Wei, K.~Lu, W.~Gan, and J.~Xue, ``Spatiotemporal features and local relationship learning for facial action unit intensity regression,'' in \emph{2021 IEEE International Conference on Image Processing (ICIP)}, 2021, pp. 1109--1113.

\end{thebibliography}
\end{document}